\documentclass[11pt]{article}

\usepackage[preprint]{acl}

\usepackage{times}
\usepackage{latexsym}

\usepackage[T1]{fontenc}

\usepackage[utf8]{inputenc}

\usepackage{microtype}

\usepackage{inconsolata}

\usepackage{graphicx}
\graphicspath{{images/}}
\usepackage{booktabs}
\usepackage{amssymb}
\usepackage{amsmath}
\usepackage{enumitem}
\usepackage{xcolor}
\usepackage{url}

\title{VISTA: A Controllable Platform for Generating and Auditing\\
Egocentric Assistance Scenarios}

\author{Yu-Hsiang Liu, Yu-Chien Tang, An-Zi Yen\\
       Department of Computer Science, National Yang Ming Chiao Tung University, Taiwan \\
      \texttt{ivesliu.ee10@nycu.edu.tw},
      \texttt{tommytyc.cs10@nycu.edu.tw},
      \texttt{azyen@nycu.edu.tw}}

\begin{document}
\maketitle
\begin{abstract}
Evaluating whether AI agents can proactively assist humans in daily activities, ranging from routine household tasks to urgent safety-critical situations, requires diverse visual data.
However, collecting such scenarios in the real world is often difficult, costly, or unsafe, and simulation environments often lack the social commonsense needed to simulate the consequences of different actions.
In this work, we present VISTA, a controllable platform that uses a user-provided scenario seed, defined as a short natural-language description of the intended assistance situation, to generate editable plans, egocentric videos, and an auditable review trail. 
VISTA structures scenario intent around three interaction modes, including reactive, explicit proactive, and implicit proactive, and two consequence families, including safety-critical and everyday inconvenience, with no-assistance cases as controls.
Its six-stage pipeline exposes the design brief, timed event script, first-frame plan, motion plan, and video plan, allowing users to revise each artifact in natural language before explicitly authorizing media generation. 
A human evaluation shows that videos retained by the complete VISTA workflow align more closely with their scenario seeds than outputs from two one-pass baselines.
VISTA thereby makes targeted egocentric scenario generation inspectable, revisable, and empirically auditable.
\end{abstract}

\section{Introduction}
\label{sec:introduction}

Assistance in daily first-person settings is expressed in different ways.
A person may directly ask for help, mention a difficulty without requesting an intervention, or provide no verbal signal at all.
The underlying event may involve a hazard, such as touching a hot surface or using a sharp tool improperly, or a mundane inconvenience, such as misplacing an item or skipping a step in a task. 
Capturing this variation requires control over what is visible, when an event unfolds, what is said, and whether assistance is warranted.

Large egocentric datasets provide authentic recordings of daily and procedural activity~\citep{grauman2022ego4d,Wang_2023_ICCV,Ragusa_2026_WACV}, and recent work synthesizes proactive dialogue from streaming first-person video~\citep{zhang-etal-2025-proactive}.
Despite their value, these recordings capture only the specific conditions and outcomes that occurred during data collection, limiting their use in controlled scenario design. 
Consequently, researchers cannot readily modify the timing of an event, a dialogue cue, or its outcome while preserving the remaining elements of the scene.
Rare hazards also present ethical concerns because scenarios involving severe consequences should not be intentionally staged. 
This restriction inevitably leaves critical scenarios underrepresented in collected datasets, creating a need for alternative methods of data generation. 
Generative video offers a potential means of addressing these gaps. 
However, direct prompting obscures important design decisions within opaque requests, making unsuccessful outputs difficult to diagnose.

In this paper, we introduce \textbf{VISTA}, a platform for generating and auditing controlled egocentric assistance scenarios.
Rather than relying on a single prompt, VISTA formulates video generation as a structured compilation process. 
A scenario seed is transformed into inspectable artifacts describing intent, scene affordances, temporal beats, interaction mode, camera constraints, and expected media evidence.
Users may edit these artifacts directly or approve scoped natural-language proposals from the human-guided VISTA Agent before authorizing generation.
Each generated video retains explicit links to its planning history and human review record, thereby supporting traceability and systematic auditing. 

VISTA organizes assistance scenarios along two orthogonal axes.
The interaction axis comprises three modes: \textit{reactive}, \textit{explicit proactive}, and \textit{implicit proactive}.
The consequence axis distinguishes \textit{safety-critical} cases from \textit{everyday-inconvenience} cases, which constitute the non-safety consequence type.
The consequence axis is not a severity scale; it broadens coverage so that the platform does not equate useful assistance with danger.
\textit{No-assistance} controls represent normal or already-resolved scenes.
The unit of evaluation in this paper is therefore the \emph{fidelity of the authored scenario and its rendered evidence}, not the behavior of a downstream agent.

Our contributions are threefold:
\begin{itemize}[leftmargin=*]
    \item We introduce a scenario model that separates interaction mode from consequence type and includes no-assistance controls.
    \item We develop an interactive platform that exposes editable planning, rendering, provenance, review, and export stages for controllable egocentric video generation.\footnote{VISTA: \url{https://nlplab-vista-research-vista-demo.hf.space/platform}}\footnote{\url{https://youtu.be/CSvrP5ykawI}}
    \item We conduct human evaluation of 60 cases and 165 judgments, and the results show that VISTA receives a larger share of best-video votes and higher case-balanced seed-match scores than two one-pass generation baselines.
\end{itemize}

\section{Related Work}
\label{sec:related_work}

\paragraph{Egocentric assistance data.}
Ego4D established broad, naturally occurring first-person video coverage across daily activities~\citep{grauman2022ego4d}.
HoloAssist records remote instructors guiding task performers and provides conversational and intervention annotations~\citep{Wang_2023_ICCV}; Ego-EXTRA similarly captures unscripted expert--trainee interaction from the trainee's viewpoint~\citep{Ragusa_2026_WACV}.
PROASSIST synthesizes proactive assistant dialogues from annotated streaming egocentric video~\citep{zhang-etal-2025-proactive}.
These resources contribute authentic activity, dialogue, or labels.
VISTA instead supports prospective authoring: a user specifies the intended condition and can revise its visual and temporal realization before rendering.

\paragraph{Controlled video generation and evaluation.}
Video diffusion models have made text-conditioned temporal synthesis practical~\citep{ho2022videodiffusion}.
EgoVid-5M targets egocentric video generation with paired action descriptions~\citep{wang2024egovid5m}, while VBench++, T2V-CompBench, and VideoScore assess general quality, compositional faithfulness, and human-aligned video quality~\citep{huang2024vbenchpp,sun2024t2vcompbench,he-etal-2024-videoscore}.
VISTA is complementary: its target is whether a generated clip preserves a specific assistance scenario, and its structured artifacts make that target inspectable before and after generation.

\paragraph{Interactive research platforms.}
An easy-to-use research platform needs to be able to incorporate human's opinion to generate ideal testing data.
For instance, Thresh exposes configurable annotation schemas~\citep{heineman-etal-2023-thresh}, ChatHF integrates multimodal feedback into an interactive interface~\citep{li-etal-2024-chathf}, and EvalAssist supports iterative synthetic-data construction for evaluation workflows~\citep{santillan-cooper-etal-2025-synthetic}.
VISTA adopts this human-in-the-loop principle for first-person scenario authoring, with an explicit provider gate and a portable review package.
Table~\ref{tab:comparison} summarizes how VISTA combines first-person evidence, grounded dialogue, user-level scenario control, interactive revision, video generation, and an inspectable provenance trail in one platform.

\newcommand{\fullcov}{\textcolor{green!45!black}{\checkmark}}
\newcommand{\partcov}{\textcolor{orange!90!black}{\(\boldsymbol{\circ}\)}}
\newcommand{\nocov}{\textcolor{gray}{--}}

\begin{table}[t]
    \centering
    \scriptsize
    \setlength\tabcolsep{2.4pt}
    \renewcommand{\arraystretch}{1.05}
    \resizebox{\columnwidth}{!}{
    \begin{tabular}{lcccccc}
    \toprule
    \textbf{Resource} & \textbf{FP} & \textbf{Dlg.} & \textbf{Ctrl.} & \textbf{Edit} & \textbf{VGen} & \textbf{Trail} \\
    \midrule
    Ego4D~\cite{grauman2022ego4d} & \fullcov & \partcov & \nocov & \nocov & \nocov & \nocov \\
    HoloAssist~\cite{Wang_2023_ICCV} & \fullcov & \fullcov & \nocov & \nocov & \nocov & \nocov \\
    Ego-EXTRA~\cite{Ragusa_2026_WACV} & \fullcov & \fullcov & \nocov & \nocov & \nocov & \nocov \\
    PROASSIST~\cite{zhang-etal-2025-proactive} & \fullcov & \fullcov & \partcov & \nocov & \nocov & \nocov \\
    EgoVid-5M~\cite{wang2024egovid5m} & \fullcov & \nocov & \partcov & \nocov & \fullcov & \nocov \\
    \midrule
    Thresh~\cite{heineman-etal-2023-thresh} & \nocov & \nocov & \fullcov & \fullcov & \nocov & \partcov \\
    EvalAssist~\cite{santillan-cooper-etal-2025-synthetic} & \nocov & \nocov & \fullcov & \fullcov & \nocov & \partcov \\
    \midrule
    \textbf{VISTA (Ours)} & \fullcov & \fullcov & \fullcov & \fullcov & \fullcov & \fullcov \\
    \bottomrule
    \end{tabular}
    }
    \caption{\textbf{Representative data, generation, and authoring resources.}
    FP: first-person evidence; Dlg.: grounded dialogue; Ctrl.: user-level scenario control; Edit: interactive revision; VGen: video generation; Trail: inspectable provenance.
    Checks, circles, and dashes denote primary, partial/adjacent, and no primary coverage.
    The comparison concerns system capability, not dataset quality.}
    \label{tab:comparison}
\end{table}
 
\section{System Architecture}
\label{sec:architecture}

\begin{figure*}[t!]
    \centering
    \includegraphics[width=\textwidth]{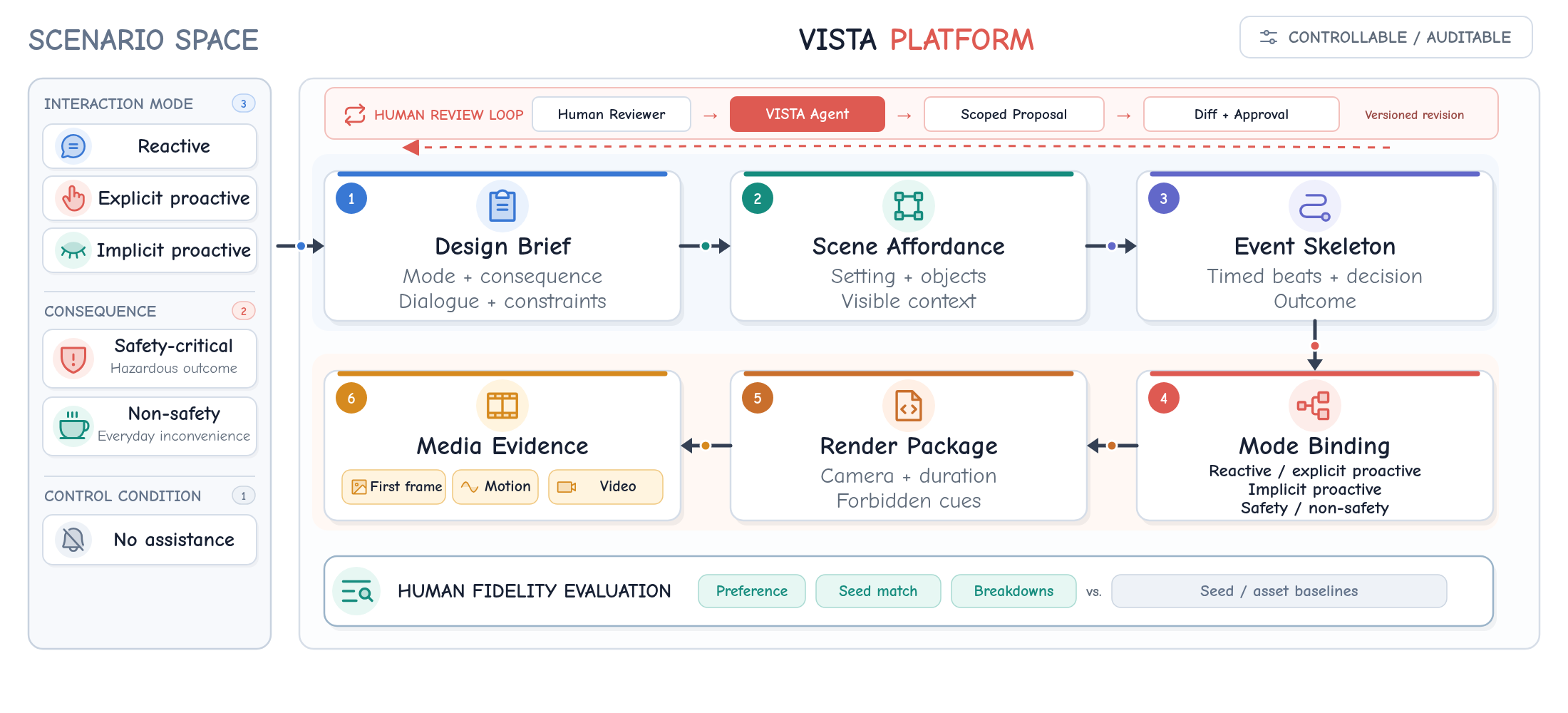}
    \caption{\textbf{VISTA platform architecture.}
    The scenario space combines three interaction modes with safety-critical and everyday-inconvenience consequences, plus no-assistance controls.
    VISTA compiles intent through six inspectable stages, routes scoped VISTA Agent proposals through human approval, and compares the resulting media with one-pass Seed+assets and Seed-only baselines using blinded preference and seed-match judgments.}
    \label{fig:architecture}
\end{figure*}

VISTA separates \emph{scenario intent} from \emph{media rendering}.
As shown in Figure~\ref{fig:architecture}, a case first receives explicit categorical and temporal structure; only then is it compiled into image, motion, and video requests.
Every stage is serializable, revisable, and linked to the final media.

\subsection{Scenario Taxonomy}
\label{sec:scenario-taxonomy}

Two orthogonal labels define an assistance case.
\textit{Interaction mode} describes how the need becomes observable; \textit{consequence type} describes the kind of outcome.
Assigning them before rendering prevents visual details from silently changing the experimental condition.

\paragraph{Interaction mode.}
In \textit{reactive} mode, the user directly asks for help.
In \textit{explicit proactive} mode, speech reveals a need or uncertainty without a request.
In \textit{implicit proactive} mode, the condition is conveyed by visual evidence and event progression rather than by a verbal signal.
These three modes cover most of the observable daily assistance support scenarios.

\paragraph{Consequence type.}
\textit{Safety-critical} type contains a plausible hazardous outcome, whereas \textit{everyday-inconvenience} type captures lower-risk mistakes, inefficiencies, or missing objects.
Including both increases semantic and visual diversity: useful assistance need not imply imminent danger.

\paragraph{No-assistance controls.}
\textit{No-assistance} controls depict normal or resolved situations.
They sit outside the assistance grid because no consequence is intended.
Their inclusion also checks that the authoring process can preserve the absence of a problem, rather than introducing a dramatic event into every generated clip.

\begin{figure*}[t!]
    \centering
    \includegraphics[width=.98\textwidth]{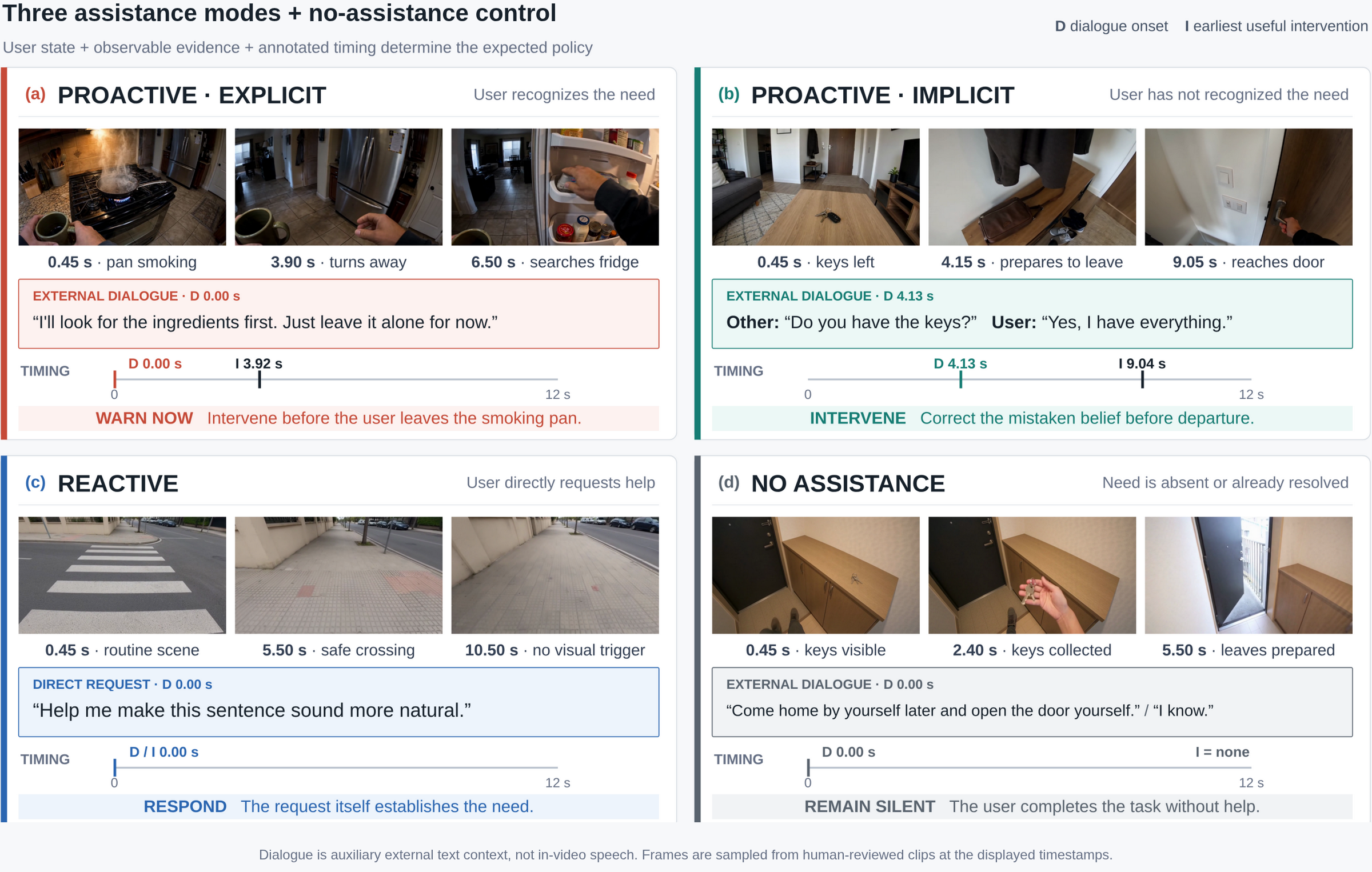}
    \caption{\textbf{Three assistance modes and one no-assistance control.}
    Each panel shows three frames from a human-reviewed clip, its auxiliary external dialogue, and the intended response policy.
    \textbf{D} marks dialogue onset and \textbf{I} the earliest useful intervention; no-assistance has no intervention target.
    Dialogue is external text context, not in-video speech.
    Safety-critical and everyday-inconvenience consequences apply orthogonally to the three assistance modes; no-assistance is a control, not a fourth assistance mode.}
    \label{fig:condition_overview}
\end{figure*}

Figure~\ref{fig:condition_overview} presents the four interaction conditions using reviewed clips, together with the auxiliary external dialogue and annotated timing evidence used to distinguish them.

\subsection{Auditable Generation Pipeline}
\label{sec:generation-pipeline}

The video generation procedure has six stages.

\begin{enumerate}[leftmargin=*, nosep]
\item The \textit{design brief} converts the seed and selected labels into a typed contract covering interaction mode, consequence type, dialogue policy, required visual evidence, and rendering constraints.
A consistency check preserves the user's selected mode and consequence.

\item \textit{Scene affordance} selects a renderable setting, task frame, object inventory, and spatial relations.
It also records the visible cues needed to identify relevant objects and states.

\item The \textit{event skeleton} expands the scene into a 12-second sequence of setup, development, decision, and outcome beats.
Each beat specifies its duration, visible goal, user action, camera behavior, required evidence, and end state.

\item \textit{Mode binding} preserves the event state and encodes the selected interaction mode through awareness, gaze, hand behavior, camera policy, dialogue affordance, and intervention window.
Reactive cases express direct help-seeking, explicit-proactive cases show recognized need or hesitation, and implicit-proactive cases keep the need unacknowledged.

\item The \textit{render package} compiles the preceding artifacts into a provider-ready plan containing duration, egocentric camera constraints, shot actions, required and forbidden cues, audio policy, and end states.
Complexity limits keep each shot focused and flag tiny interface details or small labels as fragile cues.

\item \textit{Media evidence} records the first-frame prompt and reference, optional motion sheet, video request and candidate, and provider metadata as separate artifacts.
These records connect each candidate to its plan and conditioning assets.
\end{enumerate}

\paragraph{Revision and audit.}
Artifacts form a dependency chain.
For a natural-language revision, VISTA Agent proposes a change scoped to the full chain, script, first frame, motion, or video plan; the reviewer inspects the diff and explicitly approves or discards it.
Approval records a versioned revision and its downstream invalidations before any provider execution.
The exported review package contains the current script, parsed scene, output plans, output history, and review trail.
This provenance is intended for diagnosis and reproduction; the paper's human evaluation compares only the rendered candidates and their source seed.

\section{System Interface}
\label{sec:interface}

VISTA targets multimodal researchers and evaluation designers who need to construct, inspect, and compare targeted first-person cases.
The browser interface groups the six-stage compiler into five persistent workspace views: \textit{Seed} $\rightarrow$ \textit{Intent} $\rightarrow$ \textit{Script} $\rightarrow$ \textit{Artifacts} $\rightarrow$ \textit{Trail / Export}.
Together, these views expose authoring, revision, and audit without requiring provider-specific prompts.

\paragraph{Authoring flow.}
Users select a worked example or enter a new seed.
The \textit{Intent} view presents the structured scenario specification, including the setting, objects, event, interaction mode, and consequence type, so users can correct semantic errors before temporal expansion.
The \textit{Script} view presents the scenario as an ordered sequence of timed beats, making event progression and decision points directly inspectable.
Users can target a revision to the complete scenario or a particular artifact, and VISTA identifies the downstream artifacts that require regeneration.

\begin{figure*}[t!]
    \centering
    \begin{minipage}[t]{.47\textwidth}
        \vspace{0pt}
        \centering
        \includegraphics[width=\linewidth]{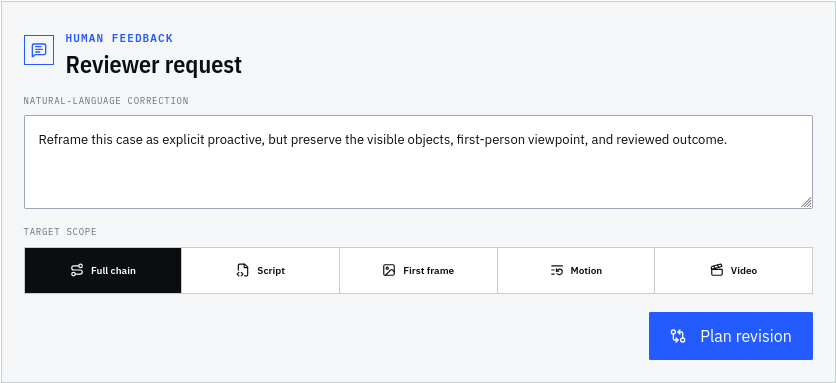}
        {\small\textbf{(a)} Reviewer feedback and target scope.\par}
    \end{minipage}
    \hfill
    \begin{minipage}[t]{.47\textwidth}
        \vspace{0pt}
        \centering
        \includegraphics[width=\linewidth]{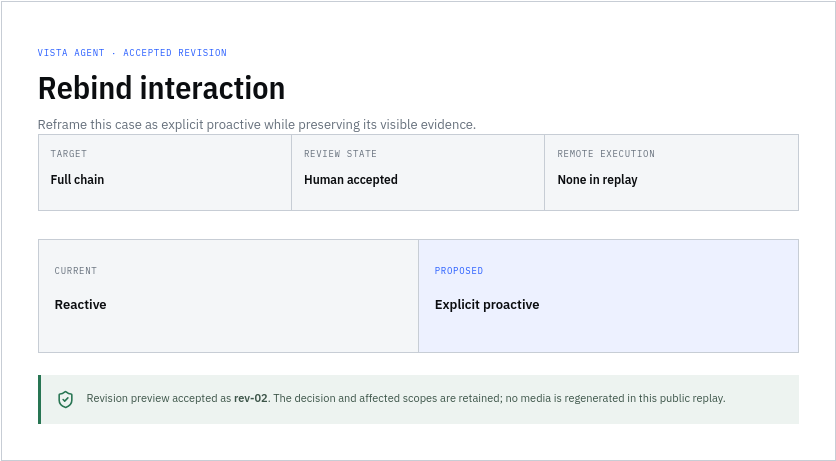}
        {\small\textbf{(b)} Human-accepted proposal and revision receipt.\par}
    \end{minipage}
    \caption{\textbf{Human-guided revision in the VISTA demo.}
    A reviewer enters a correction and selects the affected scope (a).
    VISTA Agent exposes the proposed state change and requires explicit human acceptance, which produces a versioned receipt (b).
    The hosted replay performs no remote provider call and does not regenerate source media.}
    \label{fig:vista_agent}
\end{figure*}

\paragraph{Human-guided VISTA Agent.}
Figure~\ref{fig:vista_agent} shows the revision gate in the public demo.
A reviewer supplies a natural-language correction and targets the full chain, script, first frame, motion, or video plan.
VISTA Agent returns an inspectable proposal that exposes the current and proposed state, structured actions, and downstream invalidations.
The reviewer then discards or accepts the proposal; acceptance records a versioned receipt before any provider execution.
Here, ``Agent'' denotes a human-guided revision assistant inside the platform, not a multimodal agent acting in the generated video.
In the public replay, acceptance leaves the source media unchanged and performs no remote provider call.

\paragraph{Generation and review.}
The \textit{Artifacts} view compiles request metadata for the first frame, motion sheet, and video without silently calling a media provider.
The usage bar distinguishes provider-reported values from unavailable fields, while preflight status, provider/model identity, and remote-call state remain visible.
Optional provider execution requires readiness and explicit confirmation.
Trail / Export preserves prepared requests, attempts, revisions, and provenance in a portable review package rather than only a media file.

\paragraph{Implementation and access.}
The client is implemented in React and communicates with a FastAPI service whose typed schemas validate scenario and artifact payloads.
Provider adapters are isolated behind the preparation/confirmation boundary.
The hosted preview can be explored without credentials; new provider calls require a user-supplied key and are subject to that provider's terms.\footnote{At submission time, VISTA is a hosted research preview rather than an open-source software release.}

\section{Evaluation}
\label{sec:evaluation}

We evaluate the end-to-end media-fidelity question supported by this study: do VISTA outputs preserve the user's scenario seed better?
Due to the high difficulty of automatically assessing the quality of the generated video and its alignment with the scenario seed from user, we adopt human annotation for our experiments.

\begin{figure*}[t!]
    \centering
    \includegraphics[width=.9\textwidth]{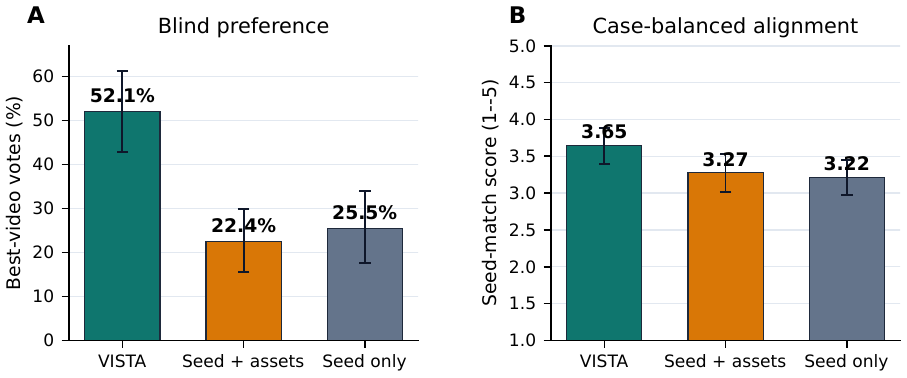}
    \caption{\textbf{Blinded human evaluation over 60 cases and 165 judgments.}
    (A) Best-video vote share.
    (B) Case-balanced mean seed-match score (1--5).
    Error bars are 95\% case-cluster bootstrap confidence intervals.}
    \label{fig:human_eval_main}
\end{figure*}

\subsection{Human Evaluation Setup}
\label{sec:human_eval_setup}

We sample 60 cases: 15 reactive, 15 explicit proactive, 15 implicit proactive, and 15 no-assistance controls.
The 45 assistance cases comprise 14 safety-critical and 31 everyday-inconvenience cases.
Each case has one video for each of three conditions.
\textbf{VISTA} is the output selected during VISTA authoring after human review.
\textbf{Seed+assets} is a one-pass baseline generated from the seed with a seed-derived first frame and, where supported, a motion sheet, but without VISTA's compiled script.
\textbf{Seed only} is a one-pass video generated from the seed and a common strict first-person instruction.
Within each case, all three conditions use the same video backend and target a 12-second, silent, first-person clip.
Explicit and implicit cases use Seedance 2.0; reactive and no-assistance cases use Sora 2.
For Sora 2, Seed+assets uses only the first-frame reference because motion-sheet conditioning is unavailable.

We invited 11 CS majoring students to contribute to the blinded case judgments.
Annotators saw the original seed and three videos labeled \textit{Video 1--3}; condition identities were hidden, and the six possible display orders were balanced across cases.
Reviewers (i) selected the video that best matched the scenario and (ii) rated each candidate's seed match on a five-point scale.
Forty-five cases received three judgments and 15 received two; we include every judgment rather than discard cases with only two judgments.

Best-video preference is the fraction of all 165 judgments selecting each condition.
For the 1--5 ratings, we first average reviewers within each case and then average the 60 case means, giving cases with two and three reviews equal weight.
We obtain 95\% confidence intervals from 20,000 case-cluster bootstrap samples, keeping all judgments for a sampled case together.
For score comparisons, we apply two-sided paired Wilcoxon signed-rank tests to case means and Holm-correct the two VISTA--baseline tests.

\subsection{Results}
\label{sec:human_eval_results}

Figure~\ref{fig:human_eval_main} shows that VISTA receives 52.1\% of best-video votes, compared with 22.4\% for Seed+assets and 25.5\% for Seed only.
The corresponding case-balanced seed-match scores are 3.65 for VISTA, 3.27 for Seed+assets, and 3.22 for Seed only.
Both VISTA--baseline score differences remain significant after Holm correction (adjusted $p=.021$ and $p=.014$, respectively).

The aggregate results support the complete workflow under the tested renderers.
Because the comparison includes VISTA's review and selection stage, it does not isolate the effect of scenario compilation alone.

Figure~\ref{fig:appendix_scenario_breakdown} reports VISTA's best-video vote share across the interaction modes and across safety-critical, everyday-inconvenience, and no-assistance groups.
The highest descriptive shares occur in implicit-proactive and everyday-inconvenience cases; because subgroup samples are small, intervals overlap, and renderer family varies by interaction mode, we interpret this breakdown as a coverage check rather than evidence of category effects.

\section{Conclusion}
\label{sec:conclusion}

In this paper, we introduce VISTA, a platform making egocentric assistance-scenario generation a controllable authoring task.
Its scenario model separates interaction mode from consequence, its compiler exposes editable intermediate artifacts, and its interface preserves a reviewable path from seed to selected media.
Experimental results show that reviewers prefer VISTA and rate them as more faithful than two one-pass baselines under the tested renderers.
We believe VISTA can effectively offer a useful platform for researchers in VLM assistants field to test in a wide range of daily scenarios, and future work will focus on stronger event-level control, improved temporal consistency, and model-agnostic generation adapters to make proactive-assistance video synthesis more reliable in safety-critical scenarios.

\section*{Limitations}
Our 60 authored cases and 11 reviewers do not exhaust daily assistance needs; review counts vary, and subgroup intervals are wide.
The VISTA condition includes authoring-stage human review and output selection, whereas both baselines are one-pass; the study therefore measures end-to-end workflow quality rather than the isolated effect of scenario compilation.
Renderer family also varies by interaction mode, precluding causal category comparisons, and provider behavior may evolve.
In addition, we do not measure physical realism, interface usability, authoring efficiency, per-constraint controllability, real-world usefulness, or downstream model behavior with other baselines; VISTA therefore cannot certify assistive-system safety.
Future work should involve a broader pool of scenario authors and represented cultures and environments, hold renderer choice fixed across interaction modes, ablate pipeline stages, and study authoring longitudinally.

\section*{Ethical Considerations}
Synthetic scenarios avoid staging dangerous incidents with real participants, but generated depictions can still contain social, cultural, or environmental bias.
We present safety cases as fictional test scenarios, not evidence about the frequency of real hazards.
No-assistance controls and everyday-inconvenience cases reduce the incentive to make every scene dramatic.
The interface separates output preparation from provider calls, requires explicit authorization, and records the provider boundary.
Users remain responsible for provider terms and for screening exported media before distribution.
The anonymized evaluation export contains case labels and judgments, but no reviewer names, email addresses, API credentials, hidden prompts, or local file paths.



\begingroup
\sloppy
\setlength{\emergencystretch}{2em}
\raggedright
\bibliography{custom}
\endgroup

\clearpage
\appendix
\section{Scenario Conditions and Evaluation Details}
\label{sec:appendix_eval}

\subsection{Human-Evaluation Coverage}

The evaluation contains 45 assistance cases and 15 no-assistance controls.
Table~\ref{tab:appendix_counts} gives the exact interaction-mode counts; the assistance subset contains 14 safety-critical and 31 everyday-inconvenience cases.
Eleven reviewers provided 165 judgments: 45 cases have three reviews and 15 have two.

\begin{table}[h]
\centering
\small
\begin{tabular}{lrr}
\toprule
\textbf{Interaction group} & \textbf{Cases} & \textbf{Judgments} \\
\midrule
Reactive & 15 & 42 \\
Explicit proactive & 15 & 41 \\
Implicit proactive & 15 & 41 \\
No assistance & 15 & 41 \\
\midrule
Total & 60 & 165 \\
\bottomrule
\end{tabular}
\caption{Human-evaluation coverage.}
\label{tab:appendix_counts}
\end{table}

\paragraph{Agreement and uncertainty.}
On the 45 fully triplicated cases, Fleiss' $\kappa=.215$, indicating modest agreement for the subjective best-video choice.
We therefore report all judgments, case-clustered intervals, and continuous seed-match ratings instead of filtering to unanimous cases.
The bootstrap resamples cases, not individual votes, and recomputes within-case means on each draw.
A case-level winner is recorded only when one condition has strictly more votes than either alternative; otherwise it is counted among the 11 split cases.

\begin{figure*}[t!]
    \centering
    \includegraphics[width=.9\textwidth]{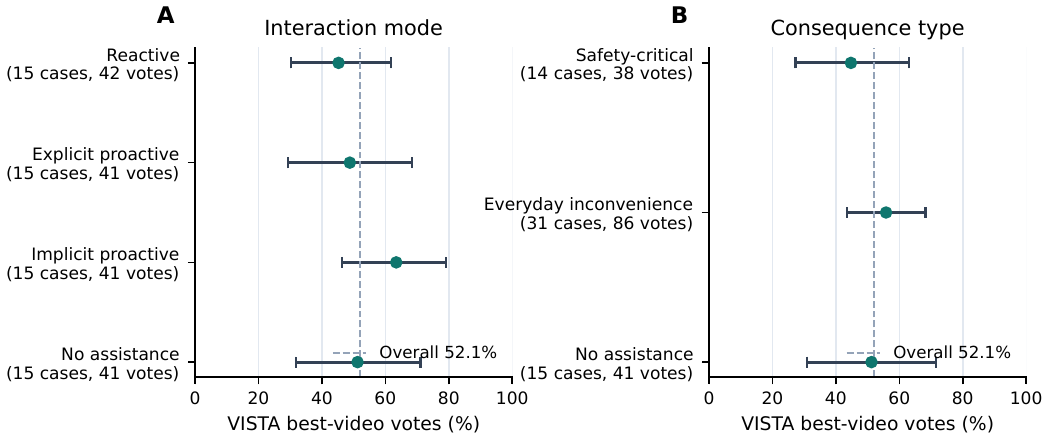}
    \caption{\textbf{Descriptive VISTA preference breakdown.}
    Dots show VISTA's best-video vote share and bars show 95\% case-cluster bootstrap intervals.
    The dashed line is the overall 52.1\% share.}
    \label{fig:appendix_scenario_breakdown}
\end{figure*}

\end{document}